# YOLO-Pose: Enhancing YOLO for Multi Person Pose Estimation Using Object Keypoint Similarity Loss

Debapriya Maji    Soyeb Nagori    Manu Mathew    Deepak Poddar
Texas Instruments Inc
{d-maji, soyeb, mathew.manu, d-poddar}@ti.com

## Abstract

*We introduce YOLO-pose, a novel heatmap-free approach for joint detection, and 2D multi-person pose estimation in an image based on the popular YOLO object detection framework. Existing heatmap based two-stage approaches are sub-optimal as they are not end-to-end trainable and training relies on a surrogate L1 loss that is not equivalent to maximizing the evaluation metric, i.e. Object Keypoint Similarity (OKS). Our framework allows us to train the model end-to-end and optimize the OKS metric itself. The proposed model learns to jointly detect bounding boxes for multiple persons and their corresponding 2D poses in a single forward pass and thus bringing in the best of both top-down and bottom-up approaches. Proposed approach doesn't require the post-processing of bottom-up approaches to group detected keypoints into a skeleton as each bounding box has an associated pose, resulting in an inherent grouping of the keypoints. Unlike top-down approaches, multiple forward passes are done away with since all persons are localized along with their pose in a single inference. YOLO-pose achieves new state-of-the-art results on COCO validation (90.2% AP50) and test-dev set (90.3% AP50), surpassing all existing bottom-up approaches in a single forward pass without flip test, multi-scale testing, or any other test time augmentation. All experiments and results reported in this paper are without any test time augmentation, unlike traditional approaches that use flip-test and multi-scale testing to boost performance. Our training codes will be made publicly available at*
https://github.com/TexasInstruments/edgeai-yolov5
https://github.com/TexasInstruments/edgeai-yolox

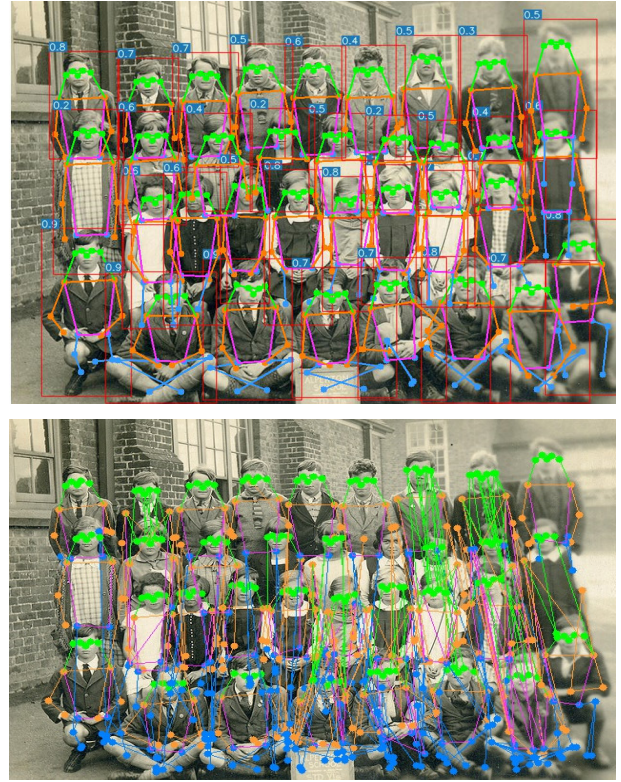

**Figure 1.** Qualitative result on a crowded COCO val2017 sample image. **Top:** Output from yolov5m6-pose shows the robustness of our approach. Keypoints of one person is never mistaken with another because of inherent grouping. **Bottom:** Output from HigherHRNetW32 showing how easily the grouping algorithm fails even when the keypoint locations are mostly correct. Bottom-up approaches are prone to such grouping errors in crowded scenes.

## 1. Introduction

Multi person 2D pose estimation is the task of understanding humans in an image. Given an input image, target is to detect each person and localize their body joints. Inferring pose of multiple person in an image can be challenging because of the variable number of persons in an image, variations in scale, occlusion of body parts, non-rigidity of the human body, and various other reasons.

Existing solutions to pose estimation are categorized into top-down and bottom-up. Top-down [30], [8], [12], [13], [19] or two-stage approaches are currently state of the art. They first employ a heavy person detector and perform single person pose estimation for each detection. The complexity of top-down approaches goes up linearly with the number of persons in an image. Most real-time applications requiring constant runtime don't prefer top-down approaches because of high complexity and variable runtime. In contrast, bottom-up [4], [5], [14], [23], [25] approaches offer constant run-time since they rely on heatmaps to detect all the keypoints in a single-shot followed by complex post-processing to group them into individual persons. Postprocessing in bottom-up approaches can involve steps like pixel-level NMS, line integral, refinement, grouping, and so on. Coordinate adjustment and refinement reduce the quantization error of down-sampled heatmaps whereas NMS is used to find local maxima in the heat-map. Even after post-processing, heat-maps may not be sharp enough to distinguish two very close joints of the same type. Again, bottom-up approaches cannot be trained end-to-end as post-processing steps are not differentiable and happen outside the convolutional network. They differ a lot across approaches, ranging from linear programming [29] to various heuristics [4]. It is difficult to accelerate them using CNN accelerators and hence are quite slow. One-shot methods [34, 35] though avoid the grouping task, do not perform at par with bottom-up approaches. They further rely on additional post-processing to boost performance.

Our motivation for this work is to solve the problem of pose estimation without heatmaps and in line with object detection since challenges in object detection are similar to that of pose estimation, e.g. scale variation, occlusion, non-rigidity of the human body, and so on. Hence, if a person detector network can tackle these challenges, it can handle pose estimation as well. E.g. latest object detection frameworks try to mitigate the problem of scale variation by making predictions at multiple scales. Here, we adopt the same strategy to predict human pose at multiple scales corresponding to each detection. Similarly, all major progress in the field of object detection are seamlessly passed on to the problem of pose estimation. Our proposed pose estimation technique can be easily integrated into any computer vision system that runs object detection with almost zero increase in compute.

We call our approach YOLO-Pose, based on the popular YOLOv5 [1] framework. This is the first focused attempt to solve the problem of 2D pose estimation without heatmap and with an emphasis on getting rid of various non-standardized post-processings that are currently used. Our approach uses the same post-processing as in object detection. We validate our models on the challenging COCO keypoint dataset [7] and demonstrate competitive accuracy in terms of AP and significant improvement in terms of AP50 over models with similar complexity.

In our approach, an anchor box or an anchor point that matches the ground truth box stores its full 2d pose along with the bounding box location. Two similar joints from different persons can be close to each other spatially. Using a heatmap, it is difficult to distinguish two spatially close similar joints from different persons. However, if these two persons are matched against different anchors, it's easy to distinguish spatially close and similar keypoints. Again, Keypoints associated with an anchor are already grouped. Hence, no need for any further grouping. In bottom-up approaches, keypoints of one person can be easily mistaken for another as shown in Figure 1 whereas our work inherently tackles this problem. Unlike, top-down approaches, the complexity of YOLO-Pose is independent of the number of persons in an image. Thus, we have the best of both top-down and bottom-up approaches: Constant run-time as well as simple post-processing.

Overall, we make the following contributions:

- We propose solving multi-person pose estimation in line with object detection since major challenges like scale variation and occlusion are common to both. Thus, taking the first step toward unifying these two fields. Our approach will directly benefit from any advancement in the field of Object detection.
- Our heatmap-free approach uses standard OD post-processing instead of complex post-processing involving Pixel level NMS, adjustment, refinement, line-integral, and various grouping algorithms. The approach is robust because of end-to-end training without independent post-processing.
- Extended the idea of IoU loss from box detection to keypoints. Object keypoint similarity (OKS) is not just used for evaluation but as a loss for training. OKS loss is scale-invariant and inherently gives different weighting to different keypoints
- We achieve SOTA AP50 with ~4x lesser compute. E.g. on coco test-dev2017, Yolov5m6-pose achieves AP50 of 89.8 at 66.3 GMACS compared to SOTA DEKR [30] achieving AP50 of 89.4 at 283.0 GMACS.
- Proposes a joint detection and pose estimation framework. Pose estimation comes almost free with an OD network.
- We propose low-complexity variants of our models that significantly outperforms real-time focused models like EfficientHRNet [15].

## 2. Related Work

Multi-person 2D pose estimation can be categorized into top-down and bottom-up approaches.

### 2.1. Top-down Methods

Top-down [8], [12], [13], [19], [20], [21] or two-stage approaches first perform human detection using a heavy person detector like Faster RCNN [21] followed by estimating 2d pose for each detected person. Hence, computational complexity increases linearly with the number of persons. Existing top-down approaches focus mostly on designing network architecture. Mask-RCNN [20] detect keypoint as segmentation mask. Simple Baseline [8] presented a simple architecture with a deep backbone and several deconvolutional layers to enlarge the resolution of the output features. These approaches are scale-invariant as they process all subjects at the same scale, achieving state-of-the-art performance on popular benchmarks. However, they are poor at handling occlusion. Our approach overcomes the challenges of occlusion to some extent as shown in Figure 3.

## 2.2. Bottom-up Methods

Bottom-up approaches [4], [5], [14], [23], [25] find out identity-free keypoints of all the persons in an image in a single shot followed by grouping them into individual person instances. Bottom-up approaches work on a probabilistic map called heatmaps that estimate the probability of each pixel containing a particular keypoint. The exact location of a keypoint is the local maxima of the heatmaps found via NMS. Bottom-up approaches are in general of lesser complexity and have the advantage of constant runtime. However, there is a substantial drop in accuracy compared to top-down approaches. There are various other steps of adjustment and refinement to extract better keypoints out of the heat-map. The part where various approaches differ is the strategy used for grouping detected body parts in an image. OpenPose [4] builds a model that contains two branches to predict keypoint heatmaps and part affinity field that are 2D vectors modeling the associations between joints. Part affinity field is used in the grouping process. In associative embedding [5], Newell et al. trains the network to predict keypoint heatmap and tag values for each joint. The loss function is defined to predict similar tag values for joints belonging to the same person and different tag values for joints belonging to different persons. Chen et al. proposed Higher HRNet [9] which uses higher output resolution to improve the precision of the prediction by a large margin. Geng et. al. [30] recently proposed disentangled keypoint representation (DEKR) that proposes a direct regression method using an offset map. It proposes a k branch structure that uses adaptive convolution to regress the offset of corresponding keypoints. This approach requires keypoint heatmap and a center heatmap for various NMS operations in post-processing. Even though the post-processing is devoid of any grouping, it is not as straightforward as ours.

## 3. YOLO-Pose

YOLO-pose is a single shot approach like other bottom-up approaches. However, it doesn't use heatmaps. Rather, it associates all keypoints of a person with anchors. It is based on YOLOv5 [1] object detection framework and can be extended to other frameworks as well. We have validated it on YOLOX [2] as well to a limited extent. Figure 2 illustrates the overall architecture with keypoint heads for pose estimation.

### 3.1. Overview

In order to showcase the full potential of our solution, we had to select an architecture that is good at detecting humans. YOLOv5 is a leading detector in terms of accuracy and complexity. Hence, we select it as our base and build on top of it. YOLOv5 mainly focuses on 80 class COCO [7] object detection with the box head predicting 85 elements per anchor. They correspond to bounding box, object score, and confidence score for 80 classes. Corresponding to each grid location, there are three anchors of different shapes.

For human pose estimation, it boils down to a single class person detection problem with each person having 17 associated keypoints, and each keypoint is again identified with a location and confidence: $\{x, y, conf\}$. So, in total there are 51 elements for 17 keypoints associated with an anchor. Therefore, for each anchor, the keypoint head predicts 51 elements and the box head predicts six elements. For an anchor with n keypoints, the overall prediction vector is defined as:

$$P_v = \{C_x, C_y, W, H, box_{conf}, class_{conf}, K_x^1, K_y^1, K_{conf}^1, \ldots \ldots \ldots, K_x^n, K_y^n, K_{conf}^n\} \quad (1)$$

Keypoint confidence is trained based on the visibility flag of that keypoint. If a keypoint is either visible or occluded, then the ground truth confidence is set to 1 else if it is outside the field of view, confidence is set to zero. During inference, we retain keypoints with confidence greater than 0.5. All other predicted keypoints are rejected. Predicted keypoint confidence is not used for evaluation. However, since the network predicts all 17 keypoints for each detection, we need to filter out keypoints that are outside the field of view. Else, there will be dangling keypoints resulting in a deformed skeleton. Existing bottom-up approaches based on heatmap don't require this since keypoints outside the field of view are not detected in the first place.

YOLO-Pose uses CSP-darknet53 [27] as backbone and PANet [6] for fusing features of various scales from the backbone. This follows by four detection heads at different scales. Finally, there are two decoupled heads for predicting boxes and keypoints.

In this work, we limit our complexity to 150 GMACS and within that, we are able to achieve competitive results. With further increase in complexity, it is possible to further bridge the gap with top-down approaches. However, we don't pursue that path as our focus is on real-time models.

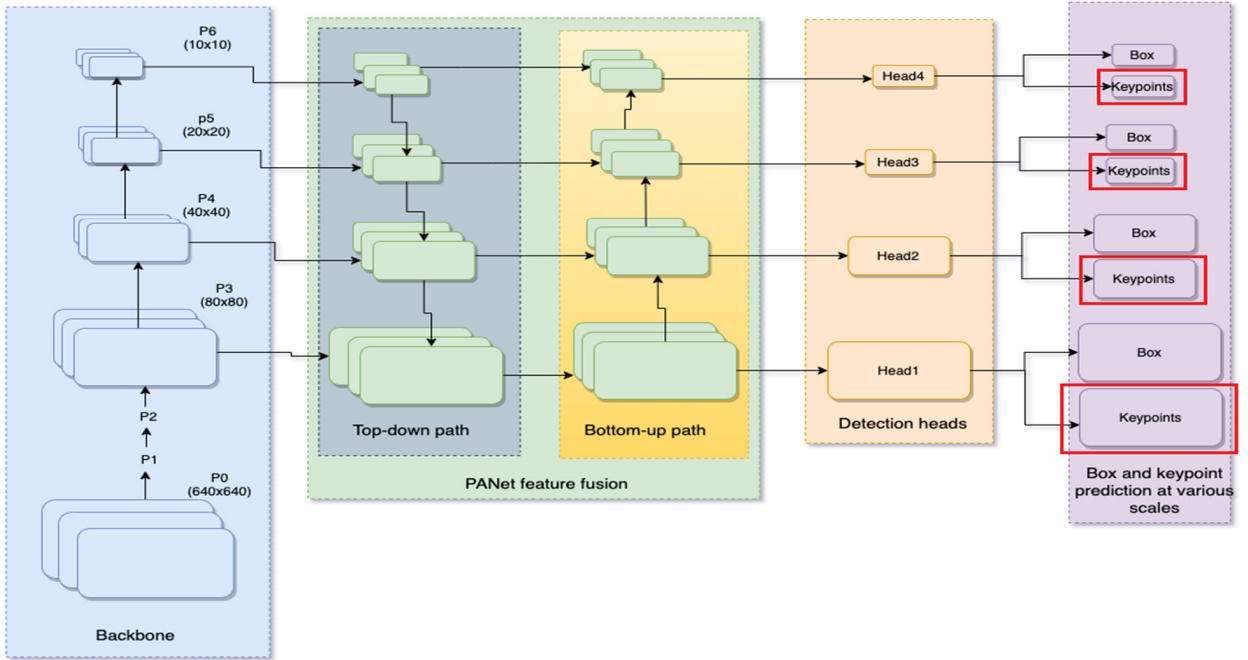

**Figure 2.** YOLO-pose architecture based on YOLOv5. Input image is passed through darknetcsp backbone that generates feature maps at various scales {P3, P4, P5, P6}. PAnet is used for fusing these feature maps across multiple scales. Output of PAnet is fed to detection heads. Finally each detection head branches into box head and keypoint head.

### 3.2. Anchor based multi-person pose formulation

For a given image, an anchor that is matched against a person stores its entire 2D pose along with bounding box. The box coordinates are transformed w.r.t the anchor center whereas box dimensions are normalized against height and width of the anchor. Similarly, keypoint locations are transformed w.r.t to anchor center. However, Keypoints aren't normalized with anchor height and width. Both Key-point and box are predicted w.r.t to the center of the anchor. Since our enhancement is independent of anchor width and height, it can be easily extended to anchor-free object detection approaches like YOLOX [2], FCOS [28].

### 3.3. IoU Based Bounding-box Loss Function

Most modern object detectors optimize advanced variants of IoU loss like GIoU [36], DIoU or CIoU [37] loss instead of distance-based loss for box detection since these losses are scale-invariant and directly optimizes the evaluation metric itself. We use CIoU loss for bounding box supervision. For a ground truth bounding box that is matched with $k^{th}$ anchor at location $(i, j)$ and scale s, loss is defined as

$$\mathcal{L}_{box}(s,i,j,k) = (1 - CIoU(Box_{gt}^{s,i,j,k}, Box_{pred}^{s,i,j,k})) \quad (2)$$

$Box_{pred}^{i,j,k}$ is the predicted box for $k^{th}$ anchor at location $(i, j)$ and scale s. In our case, there are three anchors at each location and prediction happens at four scales.

### 3.4. Human Pose Loss Function Formulation

OKS is the most popular metric for evaluating keypoints. Conventionally, heat-map based bottom-up approaches use L1 loss to detect keypoints. However, L1 loss may not necessarily be suitable to obtain optimal OKS. Again, L1 loss is naïve and doesn't take into consideration scale of an object or the type of a keypoint. Since heatmaps are probability maps, it is not possible to use OKS as a loss in pure heatmap based approaches. OKS can be used as a loss function only when we regress the keypoint location. Geng et. al. [30] uses a scale normalized L1 loss for keypoint regression which is a step towards OKS loss.

Since, we are regressing the keypoints directly w.r.t the anchor center, we can optimize the evaluation metric itself instead of a surrogate loss function. We extend the idea of IOU loss from box to keypoints. Object keypoint similarity (OKS) is treated as IOU in case of keypoints. OKS loss is inherently scale-invariant and gives more importance to certain keypoint than others. E.g. Keypoints on a person's head (eyes, nose, ears) are penalized more for the same pixel-level error than keypoints on a person's body (shoulders, knees, hips, etc.). These weighting factors are empirically chosen by the COCO authors from redundantly annotated validation images. Unlike vanilla IoU loss, that suffers from vanishing gradient for non-overlapping cases, OKS loss never plateaus. Hence, OKS loss is more similar to dIoU [37] loss.

Corresponding to each bounding box, we store the entire pose information. Hence, if a ground truth bounding box is matched with $k^{th}$ anchor at location $(i,j)$ and scale s, we predict the keypoints with respect to the center of the anchor. OKS is computed for each keypoint separately and then summed to give the final OKS loss or keypoint IOU loss.

$$\mathcal{L}_{kpts}(s,i,j,k) = 1 - \sum_{n=1}^{N_{kpts}} OKS \quad (3)$$

$$= 1 - \frac{\sum_{n=1}^{N_{kpts}} \exp\left(\frac{d_n^2}{2s^2 k_n^2}\right) \delta(v_n > 0)}{\sum_{n=1}^{N_{kpts}} \delta(v_n > 0)}$$

$d_n = $ Eucledian distance bwteen predicted and ground truth location for $n^{th}$ keypoint
$k_n = $ Keypoint specific weights
$s = $ Scale of an object
$\delta(v_n) = $ visibilty flag for each keypoint

Corresponding to each keypoint, we learn a confidence parameter that shows whether a keypoint is present for that person or not. Here, visibility flags for keypoints are used as ground truth.

$$\mathcal{L}_{kpts\_conf}(s,i,j,k) = \sum_{n=1}^{N_{kpts}} BCE(\delta(v_n > 0), p_{kpts}^n)$$

$p_{kpts}^n = $ predicted confidence for $n^{th}$ keypoint $\quad (4)$

Loss at location $(i,j)$ is valid for $K^{th}$ anchor at scale s if a ground truth bounding box is matched against that anchor. Finally, the total loss is summed over all scales, anchors and locations:

$$\mathcal{L}_{total} = \sum_{s,i,j,k} (\lambda_{cls}\mathcal{L}_{cls} + \lambda_{box}\mathcal{L}_{box} + \lambda_{kpts}\mathcal{L}_{kpts}$$
$$+ \lambda_{kpts\_conf}\mathcal{L}_{kpts\_conf}) \quad (5)$$

$\lambda_{cls} = 0.5$, $\lambda_{box} = 0.05$, $\lambda_{kpts} = 0.1$ and $\lambda_{kpts\_conf} = 0.5$ are hyper-params chosen to balance between losses at different scales.

### 3.5. Test Time Augmentations

All SOTA approaches for pose estimation rely on Test time augmentations (TTA) to boost performance. Flip test and multi-scale testing are two commonly used techniques. Flip test increases complexity by 2X whereas Multiscale testing runs inference on three scales {0.5X, 1X, 2X}, increasing complexity by (0.25X + 1X+4X) = 5.25X. With both flip test and multiscale test being on, complexity goes up by 5.25*2x=10.5X. In our table for comparison, we adjust the complexity accordingly.

Apart from an increase in compute complexity, preparing the augmented data can be expensive in itself. E.g. in flip-test, we need to flip the image that will add up to the latency of the system. Similarly, multi-scale testing requires resize operation for each scale. These operations can be quite expensive since they may not be accelerated, unlike the CNN operations. Fusing outputs from various forward passes comes with extra cost. For an embedded system, it's in best interest if we can get competitive result without any TTA. All our results are without any TTA.

### 3.6. Keypoint Outside Bounding Box

Top-down approaches perform poorly under occlusion. One of the advantages of YOLO-Pose over top-down approaches is there is no constraint for the keypoints to be inside the predicted bounding box. Hence, if keypoints lie outside the bounding box because of occlusion, they can still be correctly recognized. Whereas, in top-down approaches, if the person detection is not correct, pose estimation will fail as well. Both these challenges of occlusion and incorrect box detection are mitigated to some extent in our approach as shown in Figure 3.

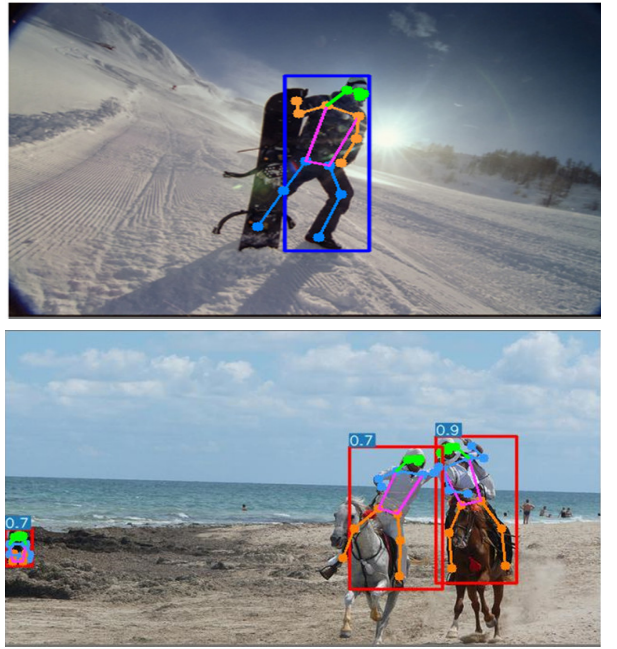

**Figure 3.** Tackling challenges of top-down approaches. **Top**: YOLO-pose correctly predicts occluded keypoint lying outside the bounding box. **Bottom:** For the person in the middle, the box detection is not exact. However, that doesn't constrain the keypoint for the right ankle to go wrong. It's predicted outside the box. Top-down approaches would have missed the keypoint for the right ankle completely.

| Method | Backbone | Input size | #params | GMACS | AP | AP50 | AP75 | AP$^L$ | AR |
|---|---|---|---|---|---|---|---|---|---|
| EfficientHRNet-H$_0$ [15] | EfficientNetB0 | 512 | 23.3M | 51.2 | 64.8 | 85.2 | 70.7 | 72.8 | 69.6 |
| HigherHRNet [9] | HRNet-W32 | 512 | 28.6M | 95.8 | 67.1 | 86.2 | 73.0 | 76.1 | - |
| HigherHRNet † [9] | HRNet-W32 | 512 | 28.6M | 502.9 | 69.9 | 87.1 | 76.0 | 77.0 | - |
| HigherHRNet [9] | HRNet-W32 | 640 | 28.6M | 149.6 | 68.5 | 87.1 | 74.7 | 75.3 | - |
| HigherHRNet † [9] | HRNet-W32 | 640 | 28.6M | 785.4 | 70.6 | 88.1 | 76.9 | 76.5 | - |
| HigherHRNet [9] | HRNet-W48 | 640 | 63.8M | 308.6 | 69.9 | 87.2 | 76.1 | 76.4 | - |
| HigherHRNet † [9] | HRNet-W48 | 640 | 63.8M | 1620.2 | **72.1** | 88.4 | **78.2** | 78.3 | - |
| DEKR[30] | HRNet-W32 | 512 | 29.6M | 90.8 | 68.0 | 86.7 | **74.5** | 77.7 | 73.0 |
| DEKR[30] | HRNet-W48 | 640 | 65.7M | 283.0 | 71.0 | 88.3 | **77.4** | 78.5 | 76.0 |
| YOLOv5s6-pose | Darknet_csp-d53-s | 960 | 15.1M | 22.8 | 63.8 | 87.6 | 69.6 | 73.1 | 70.4 |
| YOLOv5m6-pose | Darknet_csp-d53-m | 960 | 41.4M | 66.3 | 67.4 | 89.1 | 73.7 | 77.3 | 73.9 |
| YOLOv5l6-pose | Darknet_csp-d53-l | 960 | 87.0M | 145.6 | 69.4 | **90.2** | 76.1 | **79.2** | 75.9 |

† indicates multi-scale testing.
**Table 1:** Comparison with SOTA bottom-up methods on **COCO2017 val set.** Complexity for results with flip-test and multi-scale testing are adjusted with 2x and 5.25x respectively.

## 3.7. ONNX Export for Easy Deployability

ONNX [32] (Open Neural Network Exchange) is used for framework-agnostic representation of a neural network. Conversion of a deep-learning model to ONNX enables it to be efficiently deployable across various HW platforms. Post-processings in existing bottom-up approaches are not standardized and use operations that are not part of standard deep learning libraries. E.g. Bottom-up approaches based on Associative embedding use Kuhn-Munkres algorithm [29] for grouping which is not differentiable and not part of any deep-learning library.

All operators used in our model are part of standard deep learning libraries and ONNX compatible as well. Hence, the entire model can be exported to ONNX, making it easily deployable across platforms. This standalone ONNX model can be executed using ONNX Runtime [33], taking an image as input and infers bounding boxes and pose for each person in the image. No other bottom-up approaches can be exported end-to-end to an intermediate ONNX representation.

## 4. Experiments

### 4.1. COCO Keypoint Detection

**Dataset.** We evaluate our model on the COCO dataset [7]. It consists over 200,000 images with 250,000 person instances with 17 keypoints. The train2017 set includes 57K images, whereas val2017 and test-dev2017 set consists of 5K and 20K images respectively. We train the model on train2017 set and report results on both val2017 and test-dev2017 sets.

**Evaluation Metric.** We follow the standard evaluation metric and use OKS-based metrics for COCO pose estimation. We report average precision and average recall scores with different thresholds and different object sizes: AP, AP50, AP75, APL, AR.

**Training:** We adopt a similar strategy for augmentation, anchor selection, and loss weighting as YOLOv5 [1]. We use data augmentation with random scale ([0.5, 1.5]), random translation [-10, 10], random flip with probability 0.5, mosaic augmentation with probability 1, and various color augmentations.

We use SGD optimizer with a cosine scheduler. The base learning rate is set to 1e-2. We train the model for 300 epochs. Loss weights are selected to balance various losses as discussed in section 3.4.

**Testing.** We first resize the large side of the input images to the desired size maintaining the aspect ratio. The lower side of the image is padded to generate a square image. This ensures all input images have the same size.

### 4.2. Results on COCO val2017

Table 1 compares our method with other SOTA approaches based on Higher HRNet and EfficientHRNet. All our models are competitive with existing approaches of similar compute in terms of AP. Our achievements are most significant in AP50 as shown in Figure 5. E.g. YOLOv5m6-pose achieves better AP50 than SOTA DEKR[30] models with 4x higher complexity. YOLOv5l6-Pose achieves significantly better AP50 than any other bottom-up approach. Same is the case with AP$_L$. E.g. YOLOv5l6-Pose achieves better AP$_L$ than SOTA DEKR[30] models with 2x higher complexity. These results on COCO dataset positions YOLO-Pose as a strong alternative to existing state-of-the-art bottom-up approaches based on Higher HRNet.

| Method | Backbone | Input size | #params | GMACS | AP | AP50 | AP75 | AP$^L$ | AR |
|---|---|---|---|---|---|---|---|---|---|
| OpenPose [4] | - | - | - | - | 61.8 | 84.9 | 67.5 | 68.2 | 66.5 |
| Hourglass[12] | Hourglass | 512 | 277.8M | 413.8 | 56.6 | 81.8 | 61.8 | 67.0 | - |
| PersonLab [14] | ResNet-152 | 1401 | 68.7M | 911 | 66.5 | 88.0 | 72.6 | 72.3 | 71.0 |
| PiPaf [25] | - | - | - | - | 66.7 | - | - | 72.9 | - |
| HRNet [16] | HRNet-W32 | 512 | 28.5M | 77.8 | 64.1 | 86.3 | 70.4 | 73.9 | - |
| EfficientHRNet-H$_0$ [15] | EfficientNetB0 [18] | 512 | 23.3M | 51.2 | 64.0 | - | - | - | - |
| EfficientHRNet-H$_0$† [15] | EfficientNetB0 [18] | 512 | 23.3M | 268.8 | 67.1 | - | - | - | - |
| HigherHRNet [9] | HRNet-W32 | 512 | 28.6M | 95.8 | 66.4 | 87.5 | 72.8 | 74.2 | - |
| HigherHRNet [9] | HRNet-W48 | 640 | 63.8M | 308.6 | 68.4 | 88.2 | 75.1 | 74.2 | - |
| HigherHRNet † [9] | HRNet-W48 | 640 | 63.8M | 1620.2 | **70.5** | 89.3 | **77.2** | 75.8 | - |
| DEKR [30] | HRNet-W32 | 512 | 29.6M | 90.8 | 67.3 | 87.9 | 74.1 | 76.1 | 72.4 |
| DEKR [30] | HRNet-W48 | 640 | 65.7M | 283.0 | 70.0 | 89.4 | 77.3 | 76.9 | 75.4 |
| YOLOv5s6-pose | Darknet_csp-d53-s | 960 | 15.1M | 22.8 | 62.9 | 87.7 | 69.4 | 71.8 | 69.8 |
| YOLOv5m6-pose | Darknet_csp-d53-m | 960 | 41.4M | 66.3 | 66.6 | 89.8 | 73.8 | 75.2 | 73.4 |
| YOLOv5l6-pose | Darknet_csp-d53-l | 960 | 87.0M | 145.6 | 68.5 | **90.3** | 74.8 | **76.5** | 75.0 |

† indicates multi-scale testing.

**Table 2:** Comparison with bottom-up methods on **COCO2017 test-dev set.** Complexity for results with flip-test and multi-scale testing are adjusted with 2x and 5.25x respectively.

### 4.3. Ablation Study: OKS Loss vs L1 Loss.

OKS loss is one of the main contributions of our work. Since this loss is constrained unlike L1 loss, the training dynamics are much more stable. We had to tune the loss-weights a bit while training with L1 loss. OKS loss outperforms L1 loss significantly as tested on YOLv5-s6_960. We train our model with scale normalized L1 loss as in [30] to check the impact of scale invariance on our loss formulation.

| Method | LOSS | Input | AP | AP50 |
|---|---|---|---|---|
| YOLv5-s6_960 | L1 | 960 | 58.9 | 84.3 |
| YOLv5-s6_960 | Scale-normalized L1 loss | 960 | 59.7 | 84.9 |
| YOLv5-s6_960 | OKS | 960 | 63.8 | 87.6 |

**Table 3:** Comparison of various loss functions. Adding the scale information in the loss improves accuracy. However, OKS loss is the most suitable choice to obtain optimal OKS metric.

### 4.4. Ablation Study: Across Resolution.

We have trained our model across various resolutions. We chose 960 as our base resolution as we were able to get competitive performance at this resolution. Beyond this, the performance gain is saturated to a great extent as shown below for YOLOv5-s6 model:

| Method | GMACS | Input | AP | AP50 |
|---|---|---|---|---|
| YOLOv5-s6_1280 | 40.62 | 1280 | 64.9 | 88.4 |
| YOLOv5-s6_960 | 22.85 | 960 | 63.8 | 87.6 |
| YOLOv5-s6_640 | 10.15 | 640 | 57.5 | 84.3 |
| YOLOv5-s6_576 | 8.22 | 576 | 55.5 | 83.4 |
| YOLOv5-s6_512 | 6.50 | 512 | 52.3 | 80.9 |
| YOLOv5-s6_448 | 4.98 | 448 | 49.0 | 78.0 |
| YOLOv5-s6_384 | 3.65 | 384 | 44.9 | 74.2 |
| EfficientHRNet-H$_{-1}$ | 28.4 | 480 | 59.2 | 82.6 |
| EfficientHRNet-H$_{-2}$ | 15.4 | 448 | 52.9 | 80.5 |
| EfficientHRNet-H$_{-3}$ | 8.4 | 416 | 44.8 | 76.7 |
| EfficientHRNet-H$_{-4}$ | 4.2 | 384 | 35.7 | 69.6 |

**Table 4:** Comparison against SOTA low complexity (less than 30 GMACS) models on COCO2017 val dataset

At lower resolutions, YOLOv5s6-pose performs significantly better than existing state-of-the-art low complexity models like EfficientHRNet [15] on COCO val2017.

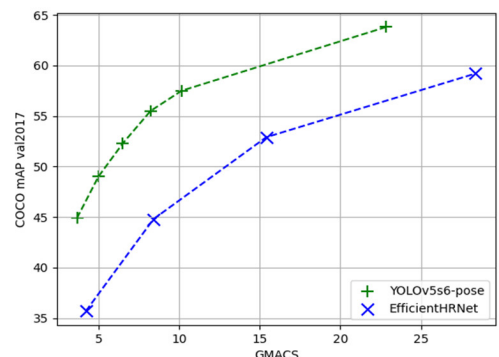

**Figure 4**. mAP vs GMACS for low complexity models on COCO val2017 (less than 30GMACS).

### 4.5. Ablation Study: Quantization

YOLOv5 models are based on sigmoid-weighted linear unit (SiLU) [39] activation. As observed by Renjie Liu et al. [38], unbounded activation functions like SiLU or hard-Swish [40] are not quantization friendly whereas models with ReLUX activations are robust to quantization because of their bounded nature. Hence, we have re-trained our models with ReLU activation. We observe ~ 1-2% drop in changing the activation from SiLU to ReLU. We call these models as YOLOv5_relu.

We have quantized these models to further position them as embedded friendly. These models can be quantized with a negligible drop (~1.2%) in accuracy as shown in the table below with the example of YOLOv5-s6_960_relu.

| Method | Bits | Input | AP | AP50 |
|---|---|---|---|---|
| YOLOv5-s6_960_relu | 32 | 960 | 61.9 | 86.7 |
| YOLOv5-s6_960_relu | 16 | 960 | 61.8 | 86.7 |
| YOLOv5-s6_960_relu | 8 | 960 | 55.6 | 81.1 |
| YOLOv5-s6_960_relu | Mixed-precision | 960 | 60.6 | 85.4 |

**Table 5:** Accuracy of our model under various quantization schemes.

With 16-bit quantization, there is a negligible drop of 0.1% in AP and no drop in AP50. With 8-bit quantization, there is a significant drop in accuracy. However, with mixed-precision, where only a selected few convolution layers are in 16 bits and the rest of the layers in 8 bits, we are able to bridge the drop from float considerably to 1.2%. In our mixed-precision setting, we had to set ~30% of all the layers in 16 bits. Since most of the CNN accelerators are optimized for integer operations, achieving good accuracy with 8-bit quantization or mixed-precision is crucial for deploying any model in an embedded system. Quantized models result in lower latency and lesser power consumption.

The above results are with Post Training Quantization (PTQ) and not Quantization aware training (QAT).

### 4.6. Results on COCO test-dev2017.

Results on test-dev dataset are presented in Table 2. Our models are competitive on the mAP metric compared to SOTA bottom-up approaches of similar compute. YOLO-Pose shows a strong result for AP50, surpassing all bottom-up approaches as shown in Figure 6. This can be attributed to the end-to-end training with an inherent grouping of our approach which is much more robust than independent grouping in other bottom-up approaches as exemplified in Figure 1. YOLO-Pose has the benefit of joint learning. Since our model can localize each person instance correctly, it's able to predict the associated keypoints with reasonably good accuracy, resulting in significant improvement in AP50. E.g. YOLOv5m6-Pose achieves an AP50 of 89.8. Whereas, DEKR with HRNetW48 backbone achieves AP50 of 89.4. In terms of compute, YOLOv5m6-Pose surpasses models with 4X more complexity in terms of AP50. Figure 5 and Figure 6 provide an easy comparison for AP50.

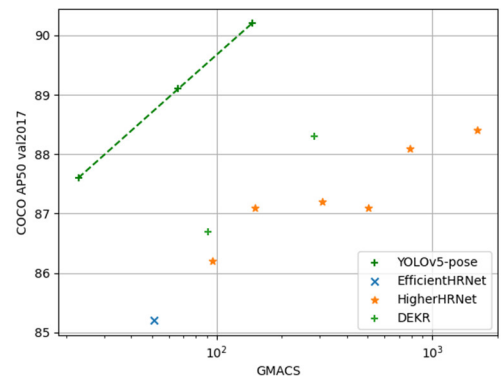

**Figure 5.** AP50 vs GMACS on COCO val2017

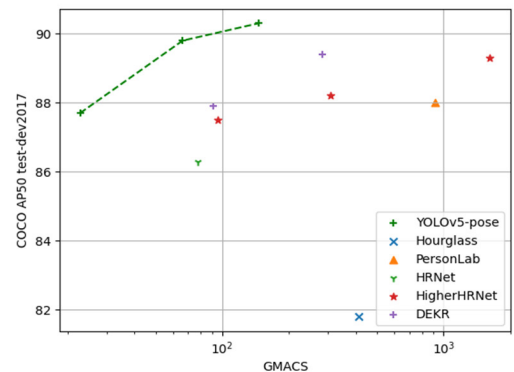

**Figure 6**. AP50 vs GMACS on COCO test-dev2017

## 5. Conclusion

We have presented an end-to-end YOLOv5 based joint detection and multi person pose estimation framework. We have shown that our models outperform existing bottom-up approaches at significantly lower complexity. Our work is the first step toward unifying the field of Object detection and human pose estimation. Until now, most of the progress in Pose estimation has happened independently as a different problem. We believe, our SOTA results will further encourage the research community to explore the potential of jointly solving these two tasks. The main motivation behind our work is to pass on all the benefits of object detection to human pose estimation since we are witnessing rapid progress in the field of object detection. We have done preliminary experimentation on extending this approach for YOLOX Object detection framework and have achieved promising results. We will extend this idea to other Object detection frameworks as well and further push the limit of efficient human pose estimation.